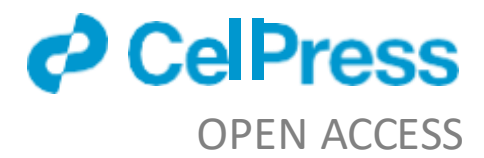

Protocol

# Protocol for evaluating ChatGPT in biomedical association generation and verification using a RAG-enabled, cross-model majority voting workflow

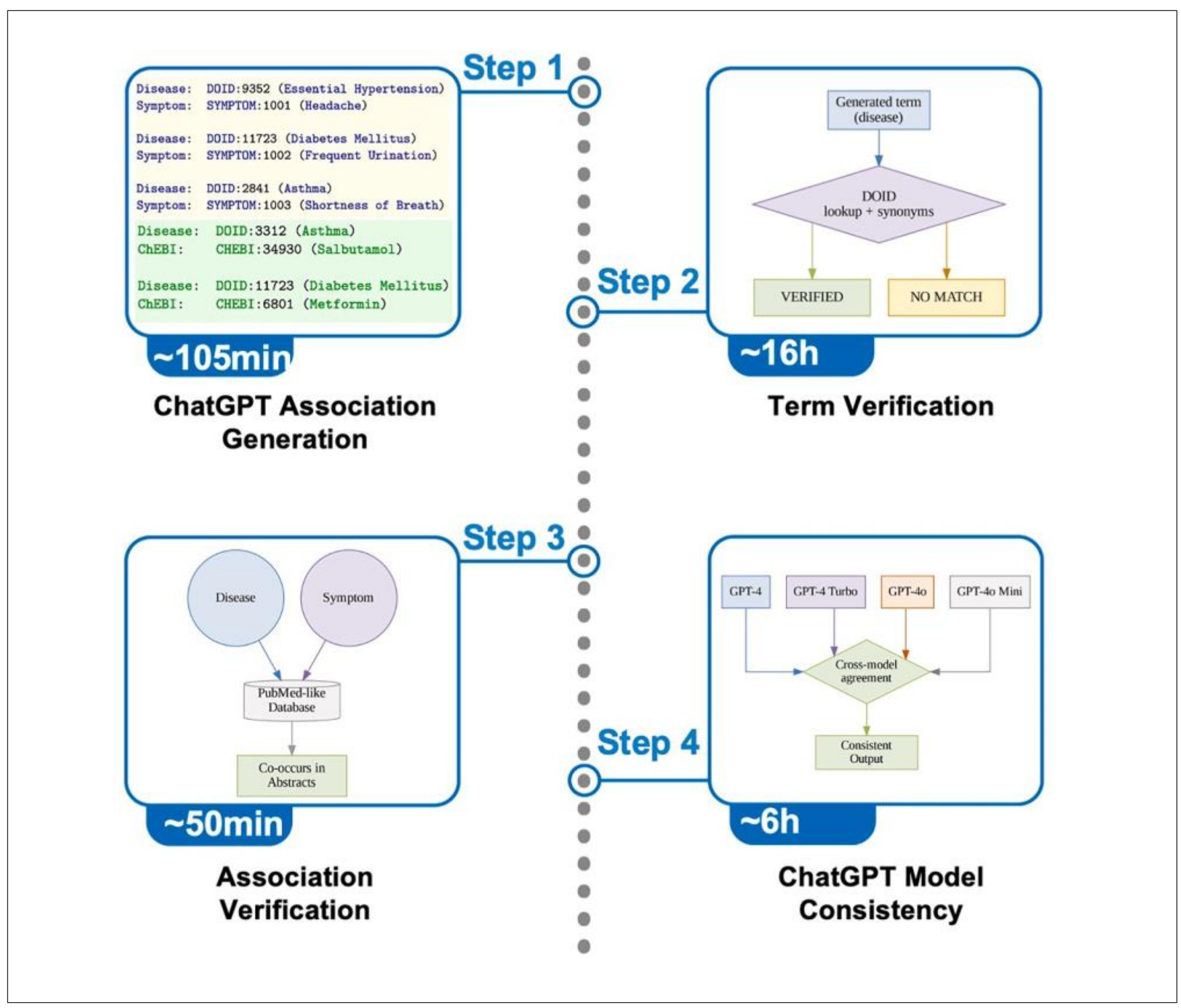

Ahmed Abdeen Hamed, Luis M. Rocha

ahamed4@nebraska.edu (A.A.H.)
rocha@binghamton.edu (L.M.R.)

Highlights

Instructions for generating and validating disease-centric JSON associations

Steps for verifying terms and associations using ontologies and literature

Guidance for performing self-consistency across multiple ChatGPT models

Procedures for building a RAG-enabled workflow for semantic term verification

 

We present a protocol to evaluate ChatGPT's ability to generate disease-centric biomedical associations. It outlines how we generate the associations, validate the biological entities using biomedical ontologies, and verify associations using literature. The protocol includes a self-consistency strategy to assess generative reliability across ChatGPT models. To address ontology exact-match limitations, we provide a use case performing semantic verification through a RAG-enabled workflow powered by open-source large language models (LLMs). This enables LLMs to establish truth over content generated by other LLMs and expose hallucination.

Publisher's note: Undertaking any experimental protocol requires adherence to local institutional guidelines for laboratory safety and ethics.

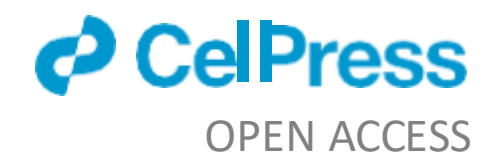

Protocol

# Protocol for evaluating ChatGPT in biomedical association generation and verification using a RAG-enabled, cross-model majority voting workflow

Ahmed Abdeen Hamed[1,2,5,6,*] and Luis M. Rocha[2,3,4,*]

[1]Department of Biochemistry, University of Nebraska-Lincoln, Lincoln, NE 68588, USA

[2]School of Systems Science & Industrial Engineering, Binghamton University, Binghamton, NY 13902, USA

[3]Universidade Católica Portuguesa, Católica Biomedical Research Centre, Lisbon, Portugal

[4]Senior author

[5]Technical contact

[6]Lead contact

*Correspondence: ahamed4@nebraska.edu (A.A.H.), rocha@binghamton.edu (L.M.R.)

## SUMMARY

**We present a protocol to evaluate ChatGPT's ability to generate disease-centric biomedical associations. It outlines how we generate the associations, validate the biological entities using biomedical ontologies, and verify associations using literature. The protocol includes a self-consistency strategy to assess generative reliability across ChatGPT models. To address ontology exact-match limitations, we provide a use case performing semantic verification through a workflow enabled by Retrieval-Augmented Generation (RAG) powered by open-source large language models (LLMs). This enables LLMs to establish truth over content generated by other LLMs and expose hallucination.**

## BEFORE YOU BEGIN

The protocol below describes the specific steps for using ChatGPT GPT-4 models for generating and verifying the factuality of disease-oriented association. However, this protocol can also permit the use of GenAI models including both commercial and open-source.

Recent advances in GenAI tools, such as ChatGPT, are influencing emerging practices in compu-tational science by enabling new forms of automation and interaction. The tool produced much excitement for the potential it offers but also raised various concerns on how various con-tent are generated, verified, or fact checked. GenAI tools in general, and ChatGPT in particular, generate responses based on user instructions through a technique known as prompt engineer-ing. That is, a user must ask a question or submit a query which triggers such tools to respond, and the outcome is immediately generated in seconds. Though impressive, such capabilities are subjects to the false fact reportings and even hallucination which is a type of answer that is completely imaginary. Since the early launch of ChatGPT and the computational science com-munity have been exploring various ways to address the generative capabilities of generating use-ful content that are subject to fact checking and verification. This task takes into account two ac-tions (1) the various ways of prompting ChatGPT and GPT-like tools, and (2) the methods of fact checking and verification of the outcome.

Prompt engineering is a new term that has emerged as ChatGPT and other GenAI tools are being launched. It refers to both the content and the style being shipped together to trigger a GenAI tool to receive a question or a query (prompts) in order to respond. While a user may ask the tool about a question such as ("what is the capital of the United Stated of America"), other advance prompts may include specific instruction, to perform a precise task, and produce a response in a given fashion and style also the user determines. For example, a user may choose to ask an open-ended question ("give me three names of over the counter drugs") this prompt is known as zero-shot,[1] which does not require any data beyond the question that is being asked. Another type of prompt where the user provides one or more example known as few-shots[2] where the examples are provided as a data for the tool to learn from. A zero-shot prompt may be transformed into few-shots if examples are provided: For the over the counter drug, one may provide advil, benadryl, Tylenol as few-shots to prompt GenAI to produce similar ex-amples. Another style is for the user to provide a context as part of the prompt and ask the tool to perform a specific task such as detecting a sentiment, extracting an entity, or identify some facts.

The task of computational fact-checking and knowledge verification has been part of our world since the widespread of the web and social media as the traditional methods have failed to address the massive scale of the content produced by such platforms.[3] Now, after the launch of GenAI tool and the massive capabilities of generating knowledge without or minimal supervision, this task be-comes essential for any tasks that utilize such tools. Prior to GenAI and ChatGPT era, computational fact checking claims was introduced to counter the spread of misinformation during political events (e.g., US election),[4,5] or global crises such as the infodemic that accom-panied the Coronavirus global pandemic.[6] When it comes to GenAI, this task requires careful consideration. On the knowl-edge part, knowledge-bases compiled from authentic sources must be compiled to offer the neces-sary ground truth for fact checking.[7,8] For example, biomedical ontology offer such mechanism that is necessary for the biomedical research as it was intro-duced in our research agenda to safeguard science authenticity.[9] On the reproducibility part, the use of traditional machine learning algorithms,[10] self-verification,[11] and also the use of one LLM against another among many other techniques may prove to be effective.[12] In the remaining part of this paper, we introduce our protocol for verifying knowledge generated from ChatGPT using biomedical ontology for name verification, and a self-consistency technique[13,14] for the knowledge generated in the form of disease-related associations.[15]

## Innovation

We evaluated the generative capabilities of ChatGPT in producing disease-centric associations. We found that LLM-generated symptoms achieved only 40–64% exact-match accuracy against the SYMPTOM ontology because the model often produced non-clinical phrasing rather than controlled vocabulary terms. Literal string matching, even with synonym expansion, could not resolve this mismatch. To address this limitation, we developed a semantic-matching workflow that integrates ontology-term embeddings, retrieval-augmented prompting, and cross-model ver-ification. All SYMPTOM ontology terms were embedded into a shared vector space, enabling retrieval of the closest neighbors. Each generated symptom was embedded at runtime, and the search was performed against the embedded ontology terms. From the search results, we re-trieved the top-30 nearest ontology terms to form a semantic context. The context was supplied to an LLM through a RAG-based prompt designed to identify the most semantically relevant ontology term while reducing hallucination by constraining the model to the context. To ensure reliability, we applied a cross-model verification strategy in which the same prompt was executed across seven diverse LLMs, and the final mapping was determined by majority vote. Using this workflow, all 10,000 generated symptoms were successfully mapped, with 76.85% supported by agreement among at least four models and the remaining 23.15% supported by at least two. This approach eliminated unmatched terms and reduced hallucination to zero in this use case, pro-viding a robust ontology-grounded semantic verification strategy. The protocol is imple-mented on the Google Cloud Platform and leverages state-of-the-art LLMs for scalable, reproduc-ible evaluation.

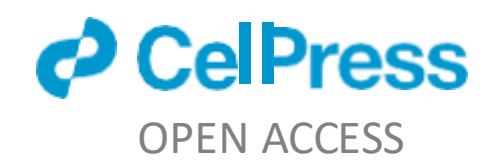

### Preparation one: ChatGPT disease-related association generation

Timing: □105 min if sequential execution generation mode is used (3 3 35 min)

This section performs the generation of disease-related associations, encodes them in JSON format, and saves them into files organized by association type: disease–gene, disease–drug, and disease–symptom.

1. Set K to 1000 for each of the disease-related associations to be generated (disease-gene, dis-ease-drug, disease-symptom).
2. Prompt-engineer for generating disease-gene associations, and store the output to a file named ("disease-gene-associations.json").
3. Prompt-engineer for generating disease-drug associations, and store the output to a file named ("disease-drug-associations.json").
4. Prompt-engineer for generating disease-symptom associations, and store the output to a file named ("disease-symptom-associations.json").

CRITICAL: The generation of valid JSON format and stored in independent files for further pars-ing and processing.

### Preparation two: ChatGPT-simulated article generation

Timing: □16.5 h if sequential execution generation mode is used

This section performs the generation of simulated PubMed-formatted articles through targeted prompt-engineering. The resulting articles serve as a testbed for evaluating the consistency of ChatGPT's knowledge, both when expressed as explicit associations and when embedded within free-text descriptions.

5. Set K to 5000 to generate disease-related simulated articles of 250-words length at mini-mum.
6. Prompt-engineer for generating simulated disease-related articles of 250-words length at mini-mum, each article has three data items (Unique ID, Title, and Abstract).
7. Store the simulated articles in a file and name it ("chatgpt-sim-articles.json").

CRITICAL: The simulated articles must have non-overlapping unique IDs to load into a py-thon dictio-nary for further processing. The abstract item must be clearly labeled as ("Abstract'').

### Preparation three: Download disease-related PubMed articles

Timing: □135 min if sequential execution generation mode is used

This section carries out the retrieval of scientific articles from the PubMed archive,[16] which are used as ground-truth material for evaluating the authenticity of the disease-related associations generated by the various ChatGPT model. Figure 1 shows the PubMed web portal with a search query "human diseases" and the search result.

8. Search the pubmed portal for disease-gene articles.
9. Specify MEDLINE as the format of the download.
10. Specify the period (2010-2014, 2015-2019, and 2020-2024).
11. Save to a file and name it for each period: Example: ("disease-gene-2014-2019.medline").
12. Repeat to produce and store articles related to the disease-symptom and disease-drug.

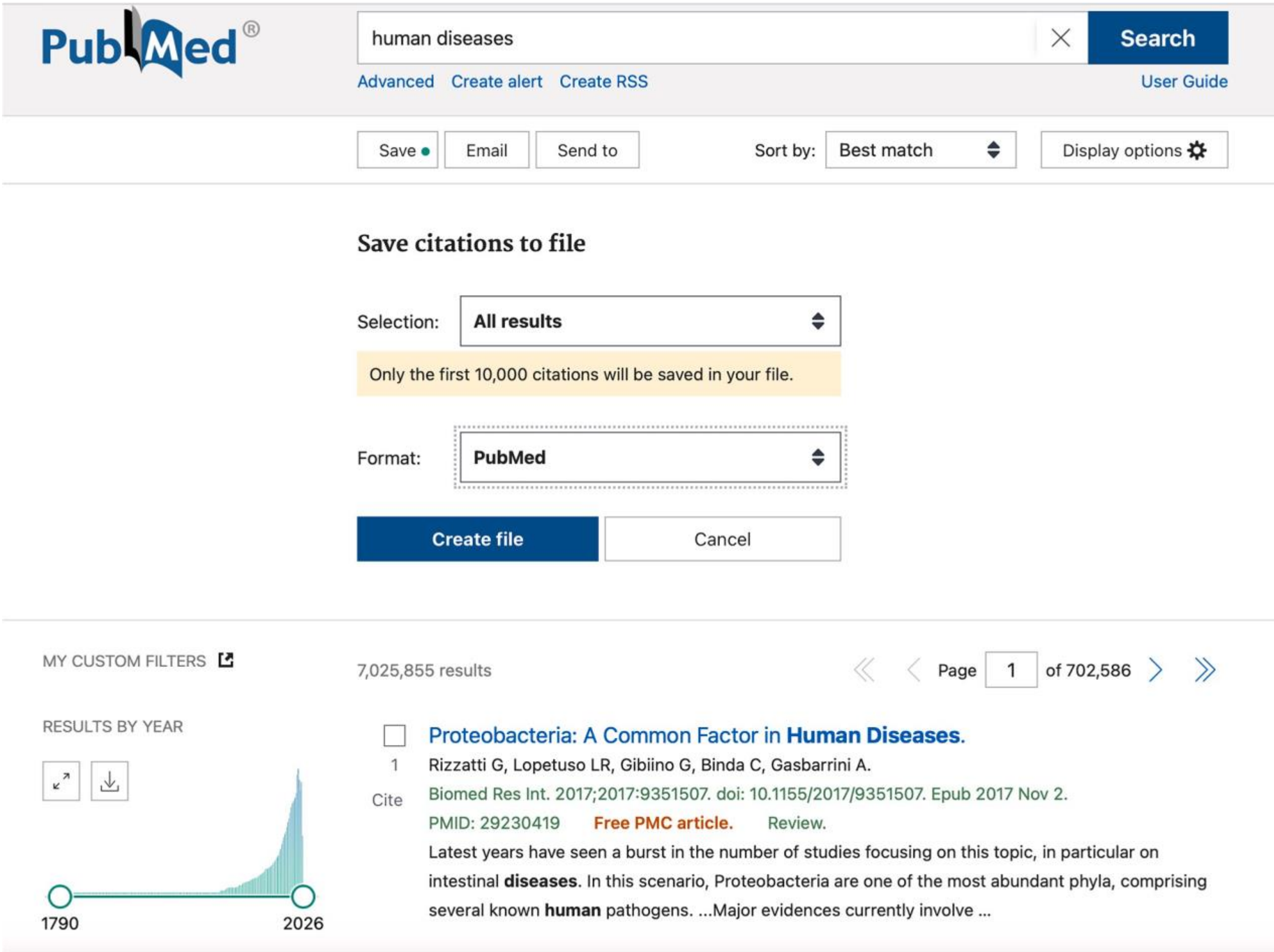

**Figure 1. For prep step 3**

A visual demonstration of querying “human diseases’’ in PubMed to retrieve articles for verifying disease-centric associations generated by ChatGPT. The illustration highlights the search query, the total number of results available (over 7 million), the web portal’s download limit (10,000 records), the required file format (PubMed), and the download mechanism, which creates a file on the user’s local system.

⚠ CRITICAL: The simulated articles must have an Abstract section and must be in a MED-LINE format.

### Preparation four: Download and process the corresponding ontologies

🕓 Timing: ~90 min if sequential execution generation mode is used

This section performs the retrieval of the relevant biomedical ontologies in the native OBO format, which serve as reference vocabularies for exact-match verification of the ChatGPT-generated terms that constitute each association.

13. Download the DOID ontology for disease name verification.
14. Download the ChEBI ontology for drug name verification.
15. Download the SYMP ontology for the symptom name verification.
16. Parse and extract the termID, term name, and synonym fields.
17. Construct a python dictionary for each ontology.
18. Save each ontology as a JSON formatted files.

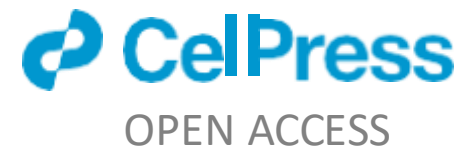

⚠ CRITICAL: The ontology format to be used for successful download is the OBO format not the OWL. Despite that the Web Ontology Language(OWL) is more expressive than OBO, we opted the use of OBO for the fact that our objective protocol was knowledge verifica-tion and not the analysis of the ontologies themselves. It was more straightforward to use the ontology as an authentic conduit of disease-related terms for the purpose of verifica-tion as the parsing of the OWL format introduces complexity that does not support the protocol of knowledge verification than OWL.

### Preparation five: Download and process gene ontology annotations

Timing: □30 min if sequential execution generation mode is used

This section converts the downloaded ontologies from OBO format into a machine-readable JSON representation. The generated JSON files contain the attributes necessary for exact-match verification.

19. Download the GOA file the genetic processes and gene names for verification.
20. Parse and extract the term IDs, names and process fields.
21. Save the annotations as JSON format for further processing CRITICAL: The gene ontology an-notations format is available in a CSV format and it must be parsed as such.

## KEY RESOURCES TABLE

| REAGENT or RESOURCE | SOURCE | IDENTIFIER |
|---|---|---|
| Software and algorithms | | |
| OpenAIModel | | |
| ChatGPT | GPT-4 APIs | https://platform.openai.com/docs/api-reference/models/list |
| ChatGPT | GPT-4turbo APIs | https://platform.openai.com/docs/api-reference/models/list |
| ChatGPT | GPT-4o APIs | https://platform.openai.com/docs/api-reference/models/list |
| ChatGPT | GPT-4o-mini APIs | https://platform.openai.com/docs/api-reference/models/list |
| Google Cloud Platform | HPC and GPU execution - | https://cloud.google.com/ |
| vLLM GitHub | Open Source LLMs | https://github.com/vllm-project/vllm |
| HuggingFace | Archive for gated LLMs | https://huggingface.co/ |
| Source code for the al-gorithms and data of the study | GitHub and Zenodo | https://doi.org/10.5281/zenodo.15199189 |

## STEP-BY-STEP METHODS DETAILS

### Major step 1: Generating disease-centric associations via ChatGPT (GPT-4)

Timing: □30 min if sequential execution generation mode is used

This section describes an algorithm for generating disease-centric associations using ChatGPT (GPT-4). The goal is to produce structured disease–symptom associations in JSON format that can be used for downstream verification and analysis.

1. Define the input parameters.
   a .Specify the number of associations to be generated (e.g., 1000).

```
Disease:  DOID:9352 (Essential Hypertension)      # high blood pressure
Symptom:  SYMPTOM:1001 (Headache)                 # head pain

Disease:  DOID:11723 (Diabetes Mellitus)          # high blood glucose
Symptom:  SYMPTOM:1002 (Frequent Urination)       # increased urination

Disease:  DOID:2841 (Asthma)                      # airway inflammation
Symptom:  SYMPTOM:1003 (Shortness of Breath)      # breathing difficulty
```

**Figure 2. For major step 1**

A ChatGPT-generated sample of DOID -SYMPTOM associa-tions. This association reflects the outcome of the prompt-engineering process performed in the same major step. In this figure, we annotate the symptom's exact match with a brief description to demonstrate that ChatGPT typically generates non-ontology phrasing, and that an exact ontology term is the exception rather than the norm.

- b .Specify the type of association to be generated (e.g., Disease-Symptom).
- c .Specify the language model M, the one-shot example s1, and the number of associa-tions N to generate.
- d .Construct the one-shot example s1 as a simple JSON object:
  - i. Assign the disease identifier "DOID:11734" the value "Epistaxis".
  - ii. Assign the symptom identifier "SYMPTOM:1080" the value "Nosebleed".

2. Construct the prompt *P* to send to the model.
   - a .Begin with an instruction message directing the assistant to generate *N* disease-symptom pairs in structured JSON format.
   - b .Concatenate the one-shot example $s_1$ to the instruction message to form a complete prompt.
3. Send the prompt *P* to the language model.
   - a .Call the function GenRes on model M with prompt P, specifying the model version (e.g., GPT-4o).
   - b .Store the model's response in variable R.
4. Process the model's response.
   - a .Check whether the response *R* is in valid JSON format.
     - i. If valid, output the response R.
     - ii. If invalid, report an error indicating improper formatting.
5. Return the final result *R* as structured disease-symptom associations.

Figure 2 presents three example disease–symptom associations, each consisting of a generated dis-ease name, symptom name, and their corresponding ontology identifiers. The ontology IDs shown here are placeholders and have not yet been validated against the official ontology terms and iden-tifiers. Note: See Methods S1 for the full RAG-enabled semantic-matching use case 1.

### Major step 2: Exact-match association term verification

Timing: For SYMPTOM ontology terms, □8–13 h (3–5 min per term 310,000 terms)

This section aims to verify whether a given term is valid within a domain-specific ontology. It checks for an exact match with known ontology terms and their synonyms. If the term or any of its synonyms exists in the ontology, it is labeled as VERIFIED; otherwise, it is labeled as UNVERIFIED. This pro-cess ensures semantic consistency when extracting terms from generative models.

6. Initialize the verification process.

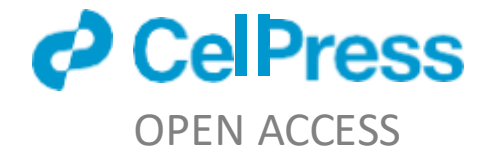

```
[
  {
    "GPT-ID": "X9B2K",
    "Title": "Targeting the BRCA1-PARP1 Axis
        in Ovarian Carcinoma",
    "Abstract": "Ovarian carcinoma remains a leading cause
        of gynecological cancer ..."
  },
  {
    "GPT-ID": "7FDJ3",
    "Title": "Interleukin-6 as a Mediator of Cardia Dysfunction
        in COVID-19",
    "Abstract": "Severe acute respiratory syndrome coronavirus
        2 (SARS-CoV-2) has..."
  },
  {
    "GPT-ID": "Q2RLP",
    "Title": "Metformin Modulates Gut Microbiota and Attenuates
        Inflammation",
    "Abstract": "Type 2 diabetes mellitus (T2DM) is a metabolic
        disorder marked..."
  }
]
```

**Figure 3. For major step 1**

Disease-centric association samples illustrating how a disease is linked to a symptom, gene, and drug, as referenced by a given biomedical ontology, DOID, SYMPTOM, GO, and ChEBI respectively.

a. Retrieve the list Lt of all terms from the domain ontology O. b
.Define the input term instance t and its expected type T.

7. Iterate through the ontology terms to verify the input. a .For
each term $t_i \longrightarrow L_t$, check if $t = t_i$.
   i. If there is an exact match, return VERIFIED.
   b .If no match is found, retrieve the list of known synonyms $L_s$ for the term $t$.
   c .For each synonym $s_j \longrightarrow L_s$, check if $s_j = t$.
   i. If a match is found among synonyms, return VERIFIED.
8. If neither the term nor its synonyms match any ontology entries. a
.Return UNVERIFIED.

Figure 3 shows a sample of the DOID ontology encoded in the OBO format (key/value pairs). The selected items are carefully picked to show the synonym key which offers the means of the semantic verification in addition to the original term name.

## Major step 3: Verification of associations using literature

Timing: ↑30–50 min for 5,000 associations when run sequentially (0.01–0.1 s per association)

This section verifies disease-symptom or drug-disease associations by checking their co-occurrence in a dataset of biomedical literature abstracts. It outputs a list of associations that appear in the same abstract, along with a computed hit ratio indicating their frequency.

9. Initialize the output list of verified associations.
   a .Create an empty list $L_v$ to store verified associations and their corresponding hit ratios.

10. For each pair of terms in the input list of associations.
   a .Set a counter to zero for tracking the number of co-occurrences in abstracts.
   b .Iterate over each abstract *B* in the dataset *D*:
      i. If both terms $p_i$ and $p_j$ appear in *B*, increment the counter by one.
   c .After scanning all abstracts, compute the hit ratio hP as the number of co-occurrences divided by the total number of abstracts.
   d .If the hit ratio is greater than zero, append the association and its hit ratio to Lv.
11. Return the list $L_v$ of verified associations and their hit ratios.

### Major step 4: Self-consistency of ChatGPT-generated associations across models

Timing: ↑2–6 h for generating 1,000 PubMed-style abstracts and performing exact-match verification across four ChatGPT models (1–8 s per API call; verification adds <2 min, sequen-tial execution)

This section describes how to use prompt engineering to generate a dataset of biomedical ab-stracts using the ChatGPT API and how to verify associations across multiple ChatGPT models. The goal is to create PubMed-style abstracts in JSON format and use them to assess whether specific biomedical associations are consistently represented across models and time periods.

12. Generate simulated articles using prompt engineering. a .Define the generation parameters.
   i. Let $n$ = 1000 be the number of simulated abstracts to generate.
   ii. Let $w$ = 250 be the desired minimum word count for each abstract.
   iii. Let $M \rightarrow \{\text{GPT-4, GPT-4turbo, GPT-4o, GPT-4o-mini}\}$ be the selected GPT model for generation.
   b .Construct the prompt.
   i. Set a system role instructing model *M* to simulate PubMed-style biomedical ab-stracts.
   ii. Set a user role with explicit instructions
   iii. Generate a list of *n* PubMed-style abstracts.
   iv. Each abstract must include three fields:
      GPT-ID (random alphanumeric string, max 5 characters),
      Title (a realistic scientific abstract title),
      Abstract (main content of *w* words).
   v. Return the abstracts in valid JSON format as an array of records.
   vi. Focus content on investigating human disease-centric associations.
   vii. Include biological details: diseases, genes, proteins, and drugs. c .Generate the dataset using model *M*.
   i. Send the prompt to model *M* via the ChatGPT API.
   ii. Parse and store the returned JSON records to a file.
   iii. Repeat as needed until the full dataset of size *n* is complete.
13. Perform exact-match verification of associations across ChatGPT models. a .Initialize parameters and counters.
   i. Let Models = {GPT-4, GPT-4turbo, GPT-4o, GPT-4o-mini}.
   ii. Let AssociationTypes = {disease-symptom, disease-gene, disease-drug}.
   iii. Let TimePeriods = {2009–2014, 2015–2019, 2020–2024}.
   iv. Initialize MatchCounter +- 0.
   b .Iterate over combinations of model, association type, and dataset period.
   c .For each model *M* in Models; association type *T* in AssociationTypes; dataset *D* in TimePeriods; abstract *t* in dataset *D*:
   i. If association *a* is exactly matched in *t*, increment MatchCounter by 1.
   ii. break the inner loop for abstract checking;
   iii. otherwise, continue to the next abstract.

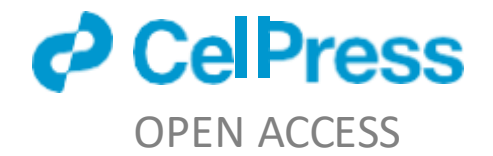

```
[
  {
    "association_type": "Disease-Symptom",
    "disease": {
      "id": "DOID:3312",
      "label": "Asthma"
    },
    "related_entity": {
      "id": "SYMPTOM:2047",
      "label": "Shortness of breath"
    }
  },
  {
    "association_type": "Disease-Gene",
    "disease": {
      "id": "DOID:9256",
      "label": "Breast carcinoma"
    },
    "related_entity": {
      "id": "GENE:675",
      "label": "BRCA1"
    }
  },
  {
    "association_type": "Disease-Drug",
    "disease": {
      "id": "DOID:9352",
      "label": "Type 2 diabetes mellitus"
    },
    "related_entity": {
      "id": "DRUG:DB00331",
      "label": "Metformin"
    }
  }
]
```

**Figure 4. For major step 2** A preview of how DOID ontology disease terms are encoded in OBO format, highlighting key -value attributes such as the synonyms field, which enables synonym-based extension of exact matching.

*Note:* This algorithm verifies whether specific biomedical associations are present in simu-lated articles generated by different ChatGPT models and uses majority voting to decide if the association is covered across time periods.

## EXPECTED OUTCOMES

### Expected outcomes for major step 1: Valid JSON associations

For each of the set of generated type of association this step produces a set of associations: (1) dis-ease-symptom, (2) disease-gene, and (3) disease-drugs. Figure 4 shows samples of each of the associations produced in one JSON format. The executor of the protocol may also choose to generate one type of association per prompt, however, for demonstration purposes, we included the three types in one listing. If the executor decides to use this format, then extra care must be taken to parse the association type before parsing and processing the association for verification.

### Expected outcomes for major step 2: Exact-match and synonym verification of association terms

The execution of this step should produce a percentage of how the entities of each association type was match against the term name and its synonyms. Since the disease terms are common among all

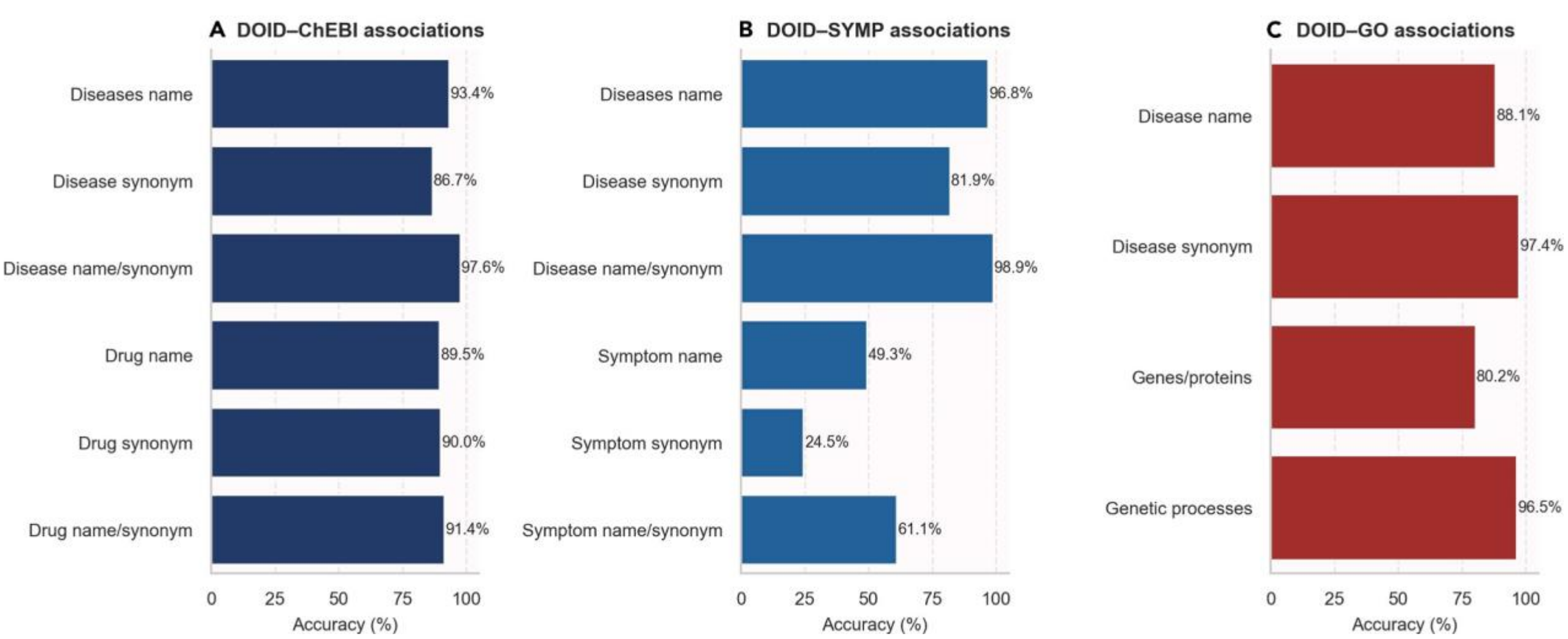

**Figure 5. For major step 2**

Exact-match accuracy across disease (DOID), chemical (ChEBI), symptom (SYMP), and gene ontology (GO) entities underlying disease-drug, disease-symptom, and disease-gene associations generated by ChatGPT in major step one.

(A) Disease-Drug Verification Result.

(B) Disease-Symptom Verification Result.

(C) Disease-Gene Verification Result.

three association types (i.e., disease-symptom, disease-drug, disease-gene), a score must be driven from each association type. Additionally, an independent score of the other entities is also expected. Our results showed that the disease terms ranged between (97% to 98%), while the drug terms were matched with a 91%. For the genetic matching, the result must also distinguishes between the terms from Gene Ontology and Gene Ontology Annotations. Two independent scores are also expected to be computed as an outcome of this step. Our study recorded 80% match for the gene names and 96% for the genetic processes. As for the synonym matching, it is expected to be observe lowest percentage when terms are exact-matched against the SYMP ontology. That is because ChatGPT is designed to interact with the general public and communicate using less formal and more social terms which are significantly different in formal language of the ontology in general and SYMPTOM ontology in particular. The outcome of this major step is shown in Figure 5. A full account of these results can also be referenced in Table 1 in Hamed et al.[15]

**Expected outcomes for major step 3: An exact-match associations query against the Pubmed-downloaded biomedical literature dataset** As the main purpose of this step is to verify whether disease-centric generated associations are valid, the expectation of this step is to produce a percentage of how many association links are covered in the biomedical literature. Since the original work tested whether the longevity of of publication (the year of publications) is a factor in the verification process, it is also expected to reproduce three sets of independent coverage one for each period: (2010-2014, 2015-2019, 2020-2024). Here are the expected coverage range for each association type: disease-drug associations (85%–90%), dis-ease-gene associations (83%–89%), gene-genetic processes asso-ciations (23%–89%), and the disease-symptom associations (49%–62%).

There are two observations that must also be expected during the execution of this step: (1) a much lower coverage for the disease-symptom association which occur as a result symptoms being described in scientific terms in literature, similar in many ways to the formal language of the SYMPTOM ontology, (2) the gene-genetic process associations shows the lowest coverage in the

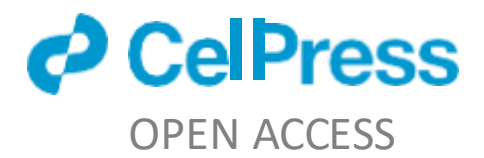

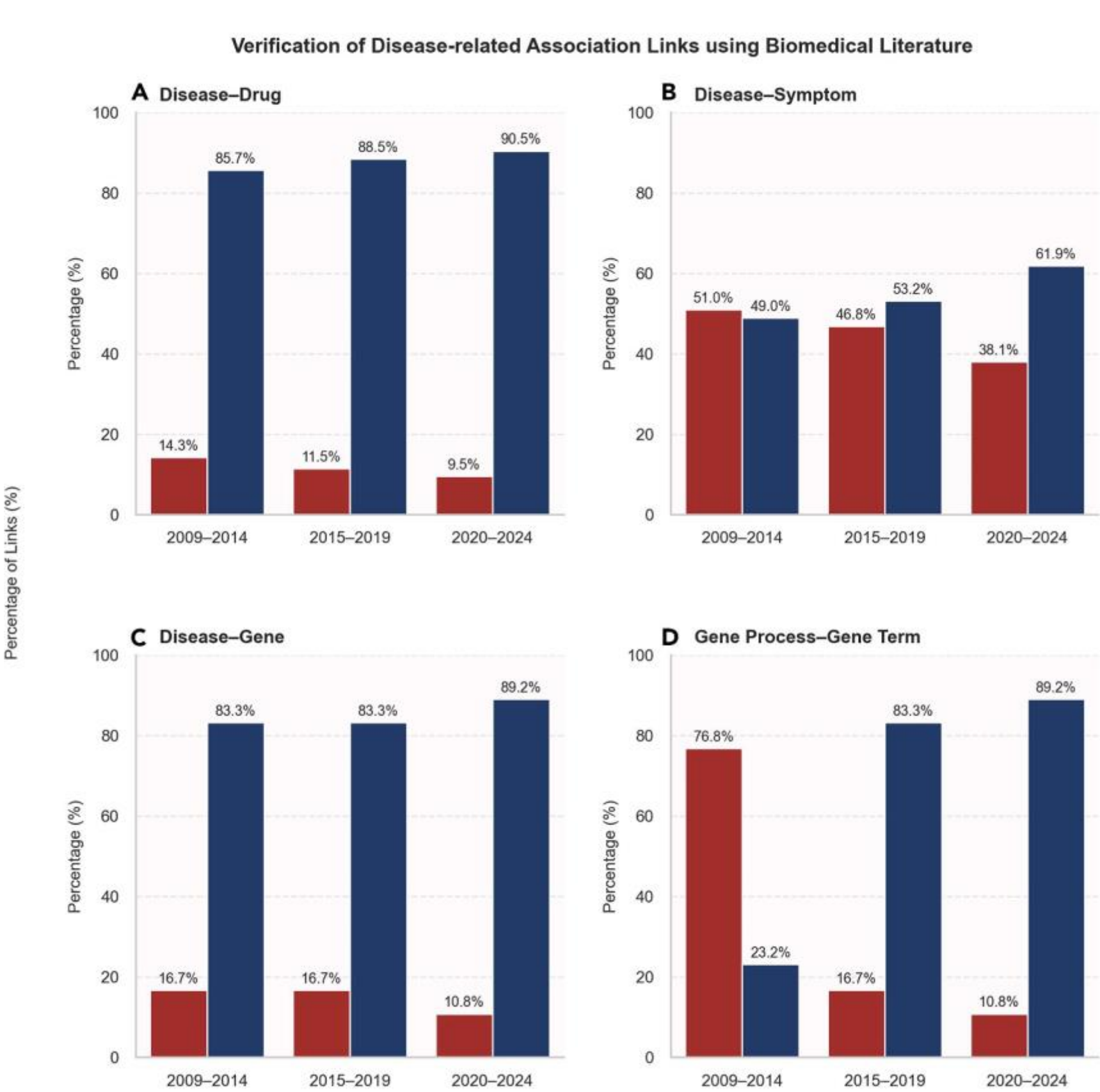

**Figure 6. For major step 3**

This step focuses on verifying the various types of associations generated by ChatGPT against biomedical literature from various periods of time (2009-2014, 2015-2019, and 2020-2024).

(A) Disease-Drug Association Link Verification.

(B) Disease-Symptom Association Link Verification.

(C) Disease-Gene Association Link Verification.

(D) Gene Process-Gene Term Correct Association.

older publications from (2009-2014), however, the same associations were covered with a much higher percentage as the year of publication progressed as they scored (83% and 89%) for the two periods (2015-2019), and (2020-2024) respectively. One interpretation of this ob-servation that ChatGPT could have not been been trained on older publication from the period (2009-2014) where such genetic knowledge was not known at the time, however, in 10-year pe-riod of time, such knowledge become available in publication as more research on genetics was performed. The outcome of the verification analysis is shown in Figure 6, while the full account of these results can also be referenced in Table 2 in Hamed et al.[15]

## Expected outcomes for major step 4, item (1): Generation of simulated articles using prompt engineering

Figure 7 shows three samples of simulated articles which were generated using the prompt engineering algorithm listed in major step 4. The figure also shows how each generated article is encoded using three specific keys: a mock-up PMID, Title, and Abstract encoded in the JSON format.

```
[
  {
    "GPT-ID": "X9B2K",
    "Title": "Targeting the BRCA1-PARP1 Axis
        in Ovarian Carcinoma",
    "Abstract": "Ovarian carcinoma remains a leading cause
        of gynecological cancer ..."
  },
  {
    "GPT-ID": "7FDJ3",
    "Title": "Interleukin-6 as a Mediator of Cardia Dysfunction
        in COVID-19",
    "Abstract": "Severe acute respiratory syndrome coronavirus
        2 (SARS-CoV-2) has..."
  },
  {
    "GPT-ID": "Q2RLP",
    "Title": "Metformin Modulates Gut Microbiota and Attenuates
        Inflammation",
    "Abstract": "Type 2 diabetes mellitus (T2DM) is a metabolic
        disorder marked..."
  }
]
```

**Figure 7. For major step 4**

A visual representation of disease-centric associations gener-ated by ChatGPT, illustrating the expected outcome of the prompt engineering also intro-duced in major step 4.

## Expected outcomes for major step 4, item (2): Exact-match associations against simulated articles generated by various ChatGPT models

The execution of this step produces a comparative analysis of how four variants of ChatGPT models provide coverage for each of the disease-centric association types. A user of this protocol should expect that newer GPT models (i.e., GPT-4o and GPT-4o-mini) are more comprehensive and provide better association coverage than older models (i.e., GPT-4 and GPT-4-turbo). Since we could only afford to generate about a thousand abstracts per association type, we had fewer chances to capture every target association. As a result, the expected coverage percentages in this step are generally lower than those observed during literature-based verification, where the volume of available con-tent is significantly larger. Importantly, lower percentages should not be interpreted as model fail-ure, but rather as a realistic outcome of sampling within constrained generative runs. What the user should look for is whether a given association appears in at least one abstract across multiple model variants, suggesting that the association is reproducibly generated and likely grounded in the model's training knowledge.

Expected coverage percentages for each model reveal that GPT-4o achieves the highest rates over-all, with approximately 14.3% coverage for disease–drug associations, 1.5% for dis-ease–gene asso-ciations, and 28.5% for disease–symptom associations. GPT-4o-mini performs next best, showing 7.7%, 3.7%, and 14.4% coverage for the same categories, respectively. Older models such as GPT-4-turbo yield more modest results—typically under 6% for disease–symptom and under 2% for the remaining types—while GPT-4 shows minimal coverage in all categories. These percentages define reasonable expectations for users aiming to replicate or benchmark the generative behavior of various GPT models under this protocol. The outcome of this verifi-cation major step is shown in Figure 8 while the full statistics are shown in Table 4 in[15] provides the full account of the expected analysis which we performed for benchmarking.

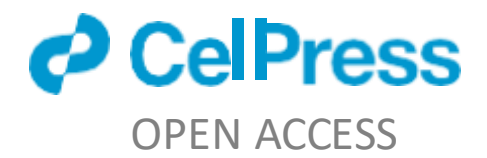

Performance of ChatGPT Variants on Biomedical Associations

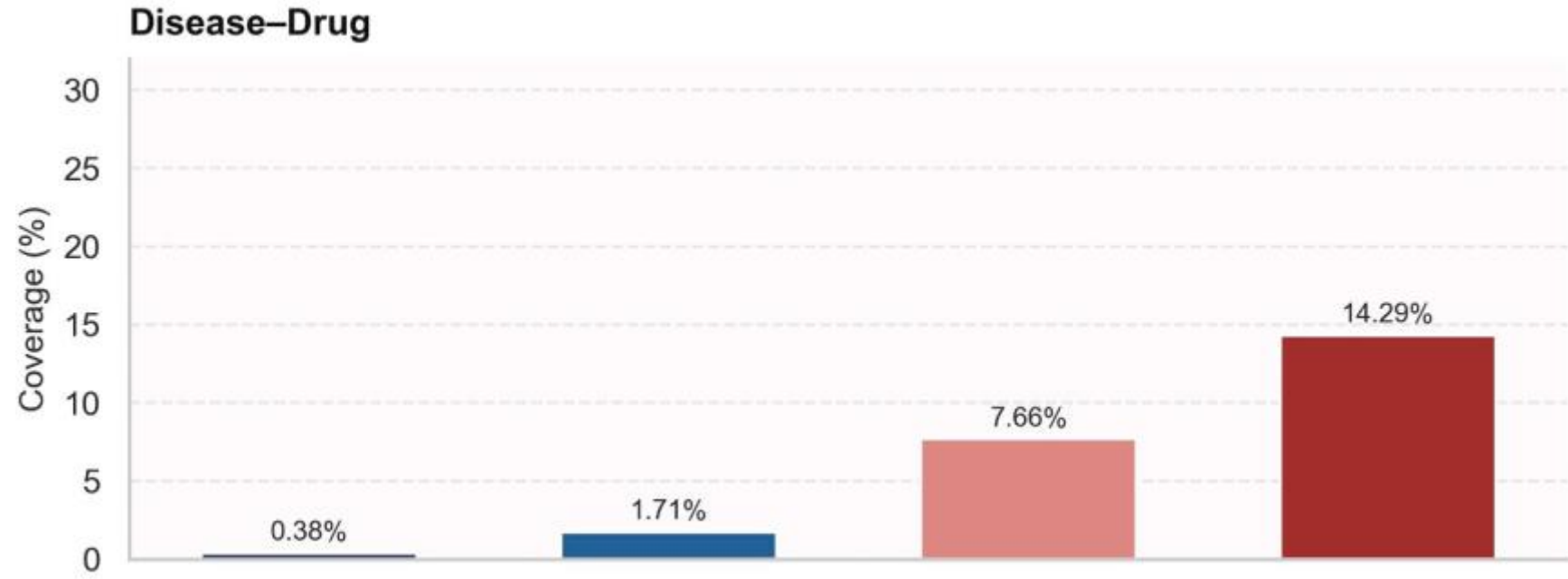

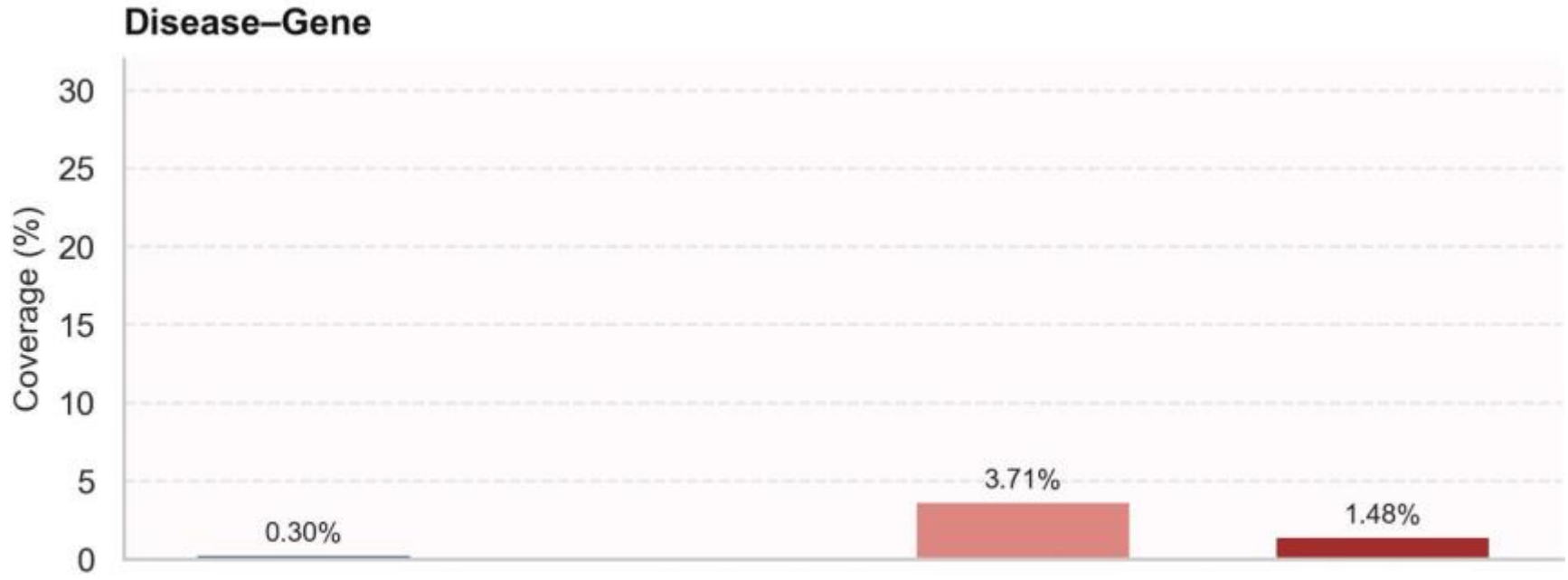

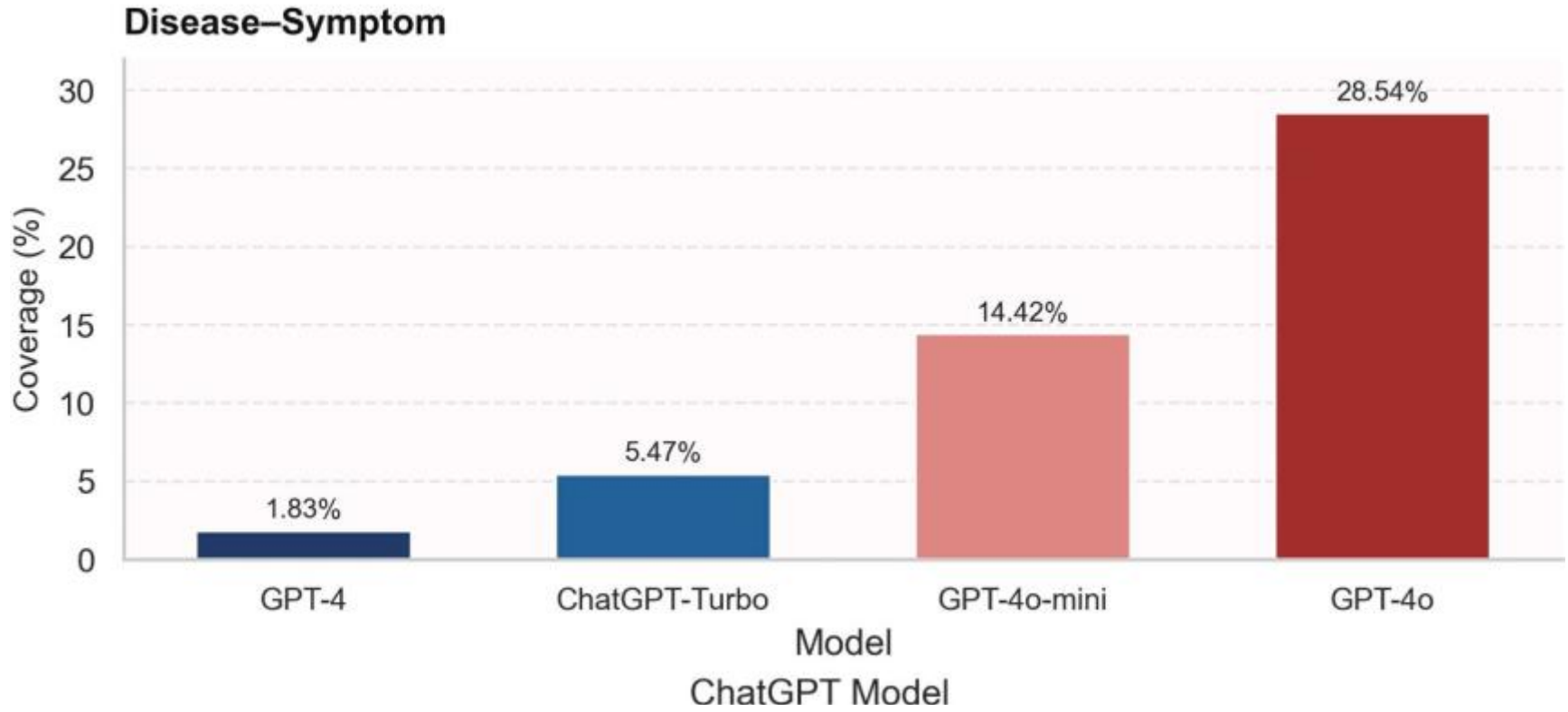

**Figure 8. For major step 4**

A visualization of association verification scores across different GPT models (GPT-4, GPT-4-turbo, GPT-4o, and GPT-4o-mini).

## LIMITATIONS

This protocol was designed to verify three core tasks, each with specific constraints that may limit the generalizability of the findings: (1) Term verification. The verification of terms was performed using formal biomedical ontologies specifically curated for disease, gene, drug, and symptom entities. The validity of the term verification outcomes may not hold if alternative lexical resources—such as general-purpose or informal dictionaries (e.g., Urban Dictionary)—are intro-duced; (2) Type of as-sociations. This study focused exclusively on verifying disease-centric associations, namely disease–symptom, disease–gene, and disease–drug relationships. Other types of associations, such as protein–protein interactions (PPI) or drug–drug interactions (DDI), were not examined and thus fall outside the validated scope of this protocol. As a result, the ap-plicability of this framework to other association types cannot be assumed without further investi-gation; (3) Causality is not supported. This study was limited to identifying the co-occurrences of terms as "associations", how-ever, it does not go deeper into whether an association could mean causality. Extra care is needed when this protocol is followed as this limitation may result in overestimating weak associations;
(4) Temporal scope of literature-based verification. The verification of associations against PubMed abstracts was limited to datasets published between 2009 and 2014. There is no guarantee that the verification outcomes would remain consistent for literature outside this temporal window, especially given shifts in terminology, knowledge, and indexing practices over time; and (5) Model-specific design. The protocol was designed and tuned specifically for use with the ChatGPT models referenced in this study. The expected re-sults may not generalize to other proprietary or open-source large language models (LLMs), as performance and behavior can vary significantly across model architectures and training data.

## TROUBLESHOOTING

### Problem 1

ChatGPT access or model incorrect configuration: Linked to major step one: prompt engineering.

### Potential solution

- Ensure the API access key is correctly placed in the environment or code. Double-check for typos or incorrect environment variable names.
- Verify that the intended GPT model (e.g., GPT-4, GPT-4o, GPT-4o-mini) is specified cor-rectly in the API call. Unsupported or deprecated model names may cause failure or silent fallback to defaults.

### Problem 2

Invalid JSON output from prompt engineering: linked to major step one: JSON association gen-eration.

### Potential solution

- The prompt may not yield well-formed JSON outputs. Validate output using a JSON linter or auto-mated schema validation before processing.
- Modify the prompt to explicitly instruct the LLM to return structured, parsable JSON using escape characters or delimiters.

### Problem 3

Ontology-specific issues: linked to major step two: semantic term verification.

### Potential solution

- Ensure the ontology is downloaded in.obo format. Other formats (e.g., OWL, JSON-LD) may not be supported by the parsing tool.
- Some ontology terms may lack a synonym field, which may affect exact match or semantic valida-tion. Consider fallback strategies such as parent-term matching or alternate identi-fiers.

### Problem 4

Association parsing issues due to non-standard tokens: linked to major steps three and four: asso-ciation extraction.

### Potential solution

- Some abstracts may use unusual tokenization or nested structures (e.g., parentheses, LaTeX formatting) that prevent accurate term matching. Consider preprocessing abstracts to normalize characters.

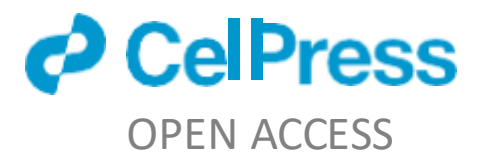

- Apply sentence segmentation or lowercasing to improve string matching success rates.

### Problem 5

PubMed file format incompatibility: linked to major step three: literature verification).

### Potential solution

- PubMed articles must be downloaded in MEDLINE format. Using other formats (e.g., XML or CSV) may break the parser.
- Some records in MEDLINE format may lack the Abstract field, which is necessary for text matching. Add a condition to skip such entries or flag them.

## RESOURCE AVAILABILITY

### Lead contact

Further information and requests for resources should be directed to and will be fulfilled by the lead contact, Ahmed Abdeen Hamed (ahamed4@unl.edu).

### Technical contact

Technical questions on executing this protocol should be directed to and will be answered by the technical contact, Ahmed Abdeen Hamed (ahamed4@unl.edu).

### Materials availability

All materials generated in this study are available for reproducibility purposes at Zenodo (DOI: https://doi.org/10.5281/zenodo.15199189).

### Data and code availability

- All original code has been deposited at Zenodo and is publicly available as of the date of publication: https://doi.org/10.5281/zenodo.15199189.
- Any additional information required to reanalyze the data reported in this article is available from the lead contact upon request.

## SUPPORTING CITATIONS

17–22

## ACKNOWLEDGMENTS

The research reported in this publication was supported in part by SUNY System Administration using the SUNY AI Platform.

## AUTHOR CONTRIBUTIONS

Conceptualization, A.A.H.; methodology, A.A.H. and L.M.R.; investigation, A.A.H. and L.M.R.; writing, A.A.H. and L.M.R.; writing – review and editing, A.A.H. and L.M.R.; resources, A.A.H.; supervision, L.M.R.

## DECLARATION OF INTERESTS

The authors declare no conflict of interest.

## DECLARATION OF GENERATIVE AI AND AI-ASSISTED TECHNOLOGIES IN THE WRITING PROCESS

During the preparation of this work, the authors used ChatGPT in order to generate associations and simulated articles to produce the datasets of this work. The use case used several open-source LLMs (including Gemma, Phi, Mistral, Qwen, among other tools). The computation was performed using Google Cloud Platform Vertex AI Workbench. After using these tools and services, the lead/technical author reviewed and edited the content as needed and takes full responsi-bility for the content of the published article.

## SUPPLEMENTAL INFORMATION

Supplemental information can be found online at https://doi.org/10.1016/j.xpro.2026.104533.

**Methods S1: A Use-Case for Performing Semantic Matching via a RAG-Enabled Workflow Empowered by Open-Source LLMs – Related to Step 1**

# Motivation

**Background:** The original protocol by Hamed et al.[15] relied on literal matching to align LLM-generated terms with biomedical ontologies, a strategy that proved insufficient when applied to symptom data. **Gap:** ChatGPT frequently produced non-clinical or colloquial symptom descriptions, which often lacked a direct lexical counterpart in the SYMPTOM ontology and resulted in substantial matching deficits. **Implication:** This limitation highlighted the need for a workflow capable of grounding generated terms semantically rather than relying on exact string equivalence. **Proposed solution:** To address this gap, we developed a semantic-matching approach that embeds both generated symptoms and ontology terms into a shared vector space, retrieves semantically related candidates, and applies LLM-based reasoning to identify the most appropriate match. A cross-model majority-vote verification step further mitigates correlated errors and increases confidence in the final mapping, providing a more reliable strategy for ontology-grounded symptom verification.

# Methods and Computational Approaches

In this section, we describe a RAG-driven semantic verification workflow composed of the following steps. (1) A generated term, treated as a query, is semantically matched against SYMPTOM ontology terms using the Facebook AI Similarity Search (FAISS)[17], a specialized vector database for high-dimensional similarity search. (2) The FAISS index is configured to retrieve the top 30 ontology terms that are semantically closest to the generated symptom. (3) The generated query and the retrieved top-30 candidates are incorporated into a structured prompt that instructs an LLM to identify the best semantic match. (4) To verify the correctness of these semantic matches, we apply a cross-model validation strategy in which the workflow is executed using seven independent LLMs. A majority vote (→ 4 models) is used as the primary decision rule, following well-established heuristics used in ensemble methods such as the *k*-nearest neighbor algorithm[18]. (5) For terms that received votes from at least two but fewer than four models (minority-vote cases), we perform an additional benchmarking analysis by randomly selecting a sample of 20 terms and comparing them against exact-match ontology lookups. The results of this comparison are presented in the manuscript. Figure S1 provides a visual overview of the full workflow.

## Setting up the Computational Environment

LLMs are known for being demanding of GPUs's computing power. Therefore, we used Google Cloud Platform (GCP) to be able to customize the computational power needed for the open-source LLMs available. We configured computing instances that can offer multiple GPU processing and perform the semantic verification task using multiple models. To accelerate the task using a given LLM, we utilized a framework known as as vLLM[19] that enables the execution of the task against multiple GPUs. We used an instances that supports the Jupyter Notebook to execute the experiments one module at a time. Following is the list of general requirements to set up an LLM-ready environment to successfully execute the task.

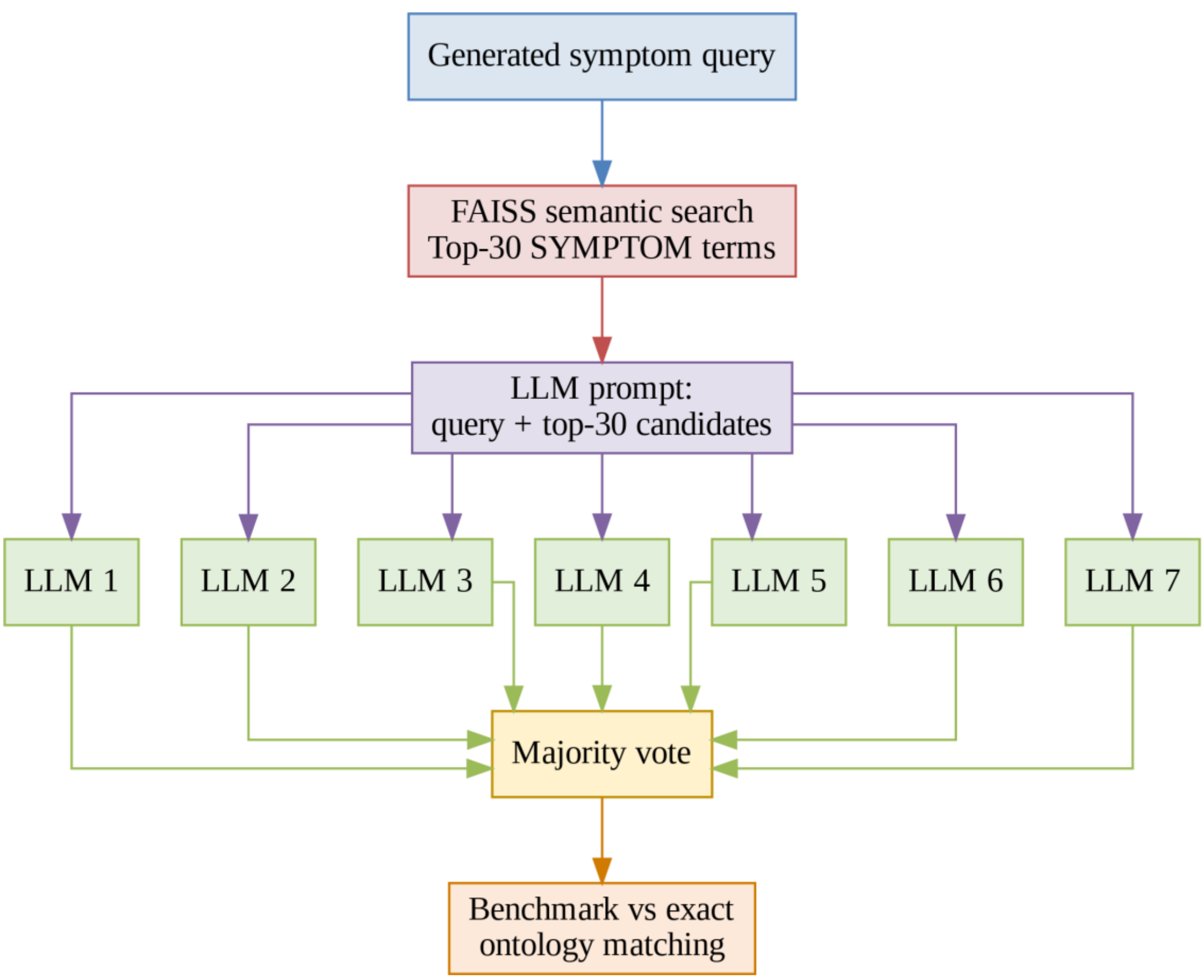

Figure S1: RAG-driven semantic verification workflow. This novel contribution illustrates FAISS-based semantic retrieval, LLM prompting, cross-model validation using seven independent LLMs, and benchmarking against exact ontology matching.

1. Securing an independent instance with at least 2 GPUs. While we used GCP, any cloud environment would offer an instance with such configurations and support for Jupyter Notebooks. 588 589 590

2. Install Python PyTorch for GPU-ready environment: 591

```
pip install torch \
--index-url https://download.pytorch.org/whl/cu121
```

3. Install the vLLM framework on the VM: 592

```
pip install vllm
```

4. Sign up into a HuggingFace account[20], and accept the term of use the open-source models used. 593 594

5. Test if the vLLM is installed properly: 595

```
python3 - << 'EOF' \
from vllm import LLM \
llm = LLM("Qwen/Qwen2.5-0.5B-Instruct")  \
```

```
print(llm.generate("Hello, vLLM!")[0].outputs[0].text) \
EOF
```

6. It is recommended to launch a terminal from within the Jupyter Notebook, launch the model of choice with the configuration being part of the excuting command (i.e., the number of GPUs, the model name, the model version, the port number running the process from a notebook):

```
CUDA_VISIBLE_DEVICES=0,1 \
vllm serve \
--model mistralai/Mistral-7B-Instruct-v0.2 \
--port 8000
```

## The Semantic Matching Using Open-Source LLMs

Since the objective of this use case is to address the issue of the unverified symptoms in the disease-symptom associations generated by an LLM, the process requires the follow two input data items:

7. A new list of 10,000 generated disease-symptom associations
8. The list of SYMPTOM ontology Term-ID and and Term.

## Executing the Retrieval-Augment generation (RAG) to Produce Top-30 Candidates

The RAG process takes a single symptom term and the full list of SYMPTOM ontology terms as input. It returns the Top-K most similar ontology terms, where K is set to 30. For each symptom, the model identifies the 30 closest ontology terms based on semantic similarity. Below, we outline the steps used to perform the RAG retrieval and the Python functions responsible for generating the Top-30 candidates shown in code snippet below.

9. Installing the necessary transformers:

```
pip install sentence transformers
```

10. Selecting an embedding model that is compatible with the open-source LLMs used:

```
intfloat/e5-large
```

11. Process a symptom term as the main query parameter.
12. Perform the RAG search which produces a list of top-30 similar candidate terms and their corresponding scores
13. Both the candidate terms and a symptom as returned as an input for a zero-shot prompt to perform the semantic matching.

```python
from sentence_transformers import SentenceTransformer
import numpy as np
import faiss

def build_ontology_index(labels):
    embed_model = SentenceTransformer("intfloat/e5-large", device="cpu")
    embeddings = embed_model.encode(labels, normalize_embeddings=True)

    dim = embeddings.shape[1]
    index = faiss.IndexFlatIP(dim)
    index.add(embeddings)

    return embed_model, index, embeddings

def retrieve_candidates(symptom, embed_model, index, labels, k=30):
    query = embed_model.encode([symptom], normalize_embeddings=True)
    scores, idx = index.search(query, k)
    return [labels[i] for i in idx[0]]
```

## The Best Semantic Match Prompt Engineering

To be able to match each of the generated symptoms, we designed a highly configurable prompt template that accepts the following parameters:

14. The ontology term generated.
15. The Top-K candidates retrieved terms produced by the Retrieval-Augmented Generation process.
16. A given open-source LLM. Here we experimented with the following models (Qwen2.5, MistralAI, Microsoft/Phi3, Google/Gemma, Meta-llama/Llamm-3.1).
17. The instructions to producing valid output in the form of JSON format to represent the term and its semantic best match.

The code below shows how the prompt is encoded and the necessary configuration parameters.

```
prompt = f"""
    Match the symptom to the closest ontology label.
    Symptom: "{symptom}"

    Candidate labels:
    {json.dumps(candidates)}

    Return ONLY:
    <json>
    {{
      "symptom": "{symptom}",
      "best_match": "<exact label from the list or NONE>"
    }}
    </json>
    """
```

It also illustrates how the model was instructed to produce a JSON-formatted output, which is essential for post-processing the terms and their semantic matches. Specifically, each semantic match is voted on by all LLMs to verify that the generated term is consistently mapped to a single SYMPTOM ontology term with a precise SYMP term ID.

## Semantic Matching of Generated Terms Against Top-30 RAG Candidates

This process performs the semantic matching of each of the 10,000 generated terms against its corresponding 30 candidates. Each term is processed individually by comparing it to its 30 candidate terms across all LLMs. This procedure was repeated for all 10,000 terms and run independently for each of the seven LLMs. Below are the computational steps used to produce a single semantic match between each generated term and its SYMP ontology term ID.

18. Load the ontology semantic labels as key/value pairs: (Term-ID and Term-Label)
19. List of all the 10,000 generated to be semantically matched one term as at time
20. Execute the RAG process for each of the generated terms and produce the 30-candidates perm
21. Configure the zero-shot prompt with the SYMP term, candidates, and the LLM
22. Extract the semantically identified term and identify its Term-ID from the ontology
23. Return a record of a valid JSON record that compile all the results:

```
JSON_record = {generated-symp, onto-best-match,
    onto-term-item, status-of-identification}
```

24. Store the JSON records as JSONL file for further verification and computing the majority vote among all the models.

## Cross-Model Verification: a Majority-vote Heuristic

Inspired by the data cross-validation and the strategy of fighting fire with fire[12], we performed a verification process known as cross-models. The process is compute the semantic match of a generated symptom uniformly using 7 independent LLMs. To reach a sound prediction of a SYMP term, we used the majority vote among all the models to determine that a generated term has been semantically resolved and mapped to one ontology term as discussed above. This process analyzed the outcome of each of the 7 LLM in the form of a JSON format model was presented as a JSONL file. The final outcome of this process was a record that represents each term, the SYMPTOM ontology term label and ID, the number of votes, and whether the vote was a majority or a minority as shown in the code below.

```
import json

files = [
    "~/../outs/semsymp-analysis-gemma-3-4b-it.jsonl",
    "~/../outs/semsymp-analysis-gemma-7b-it.jsonl",
    "~/../outs/semsymp-analysis-Llama-3.1-8B.jsonl",
    "~/../outs/semsymp-analysis-mistral-02102026.jsonl",
    "~/../outs/semsymp-analysis-phi-02102026.jsonl",
    "~/../outs/semsymp-analysis-qwen-02102026.jsonl",
    "~/../outs/semsymp-analysis-Qwen2.5-14B-Instruct.jsonl"
]

# Load all model outputs
all_models = [ [json.loads(line) for line in open(f)] for f in files  ]

majority_results = []

for i in range(len(all_models[0])):
    # Collect the best-match term from each model for symptom i
    matches = [m[i]["onto_best_match"] for m in all_models]

    # Count votes
    vote_counts = {}
    for m in matches:
        vote_counts[m] = vote_counts.get(m, 0) + 1

    # Determine majority
    best_term = max(vote_counts, key=vote_counts.get)
    votes = vote_counts[best_term]

    majority_results.append({
        "generated_symp": all_models[0][i]["generated_symptom"],
        "majority_term": best_term,
        "votes": votes,
        "all_model_matches": matches,
        "status": "majority verified" if votes >= 4 else "minority/no-consesus"
    })
```

## Verification of Minority-Vote Semantic Matches

To better characterize the behavior of terms resolved through minority voting, we randomly sampled several sets of 20 terms from the 23.15% of cases that received $\rightarrow 2$ and $\uparrow 3$ votes. Each sampled term was evaluated against the SYMPTOM ontology using a three-step procedure: (1) we performed an exact literal match against all SYMPTOM ontology labels to determine whether the generated term existed verbatim; (2) we retrieved the minority-vote term selected by the LLM ensemble; and (3) we extracted the corresponding SYMPTOM term identifier associated with that semantically matched term. The following code snippet illustrates this analysis workflow:

```

# Load SYMP ontology terms
analysis_table = []
symp_terms = load_symptom_terms("../data/symp-ontology.json")

# Build analysis table for selected minority-vote terms
for rec in selected_terms:
    generated = rec["generated_symp"].lower()
    minority_label = rec["majority_term"].lower()
    minority_id = rec["majority_term_id"]

    # 1. Literal match in SYMP ontology?
    literal_match = generated in symp_terms

    # Append structured record for reporting
    analysis_table.append({
        "generated_symptom": generated,
        "found_in_symptom_ontology": literal_match,
        "minority_vote_term": minority_label,
        "minority_vote_term_id": minority_id
    })
```

The results included in Table S1 shows a snapshot of a 20 randomly selected terms set had no exact match in the ontology. In fact, all the 5 random sets had no exact match. The table also demonstrates that such terms were indeed matched, though using a minority-vote, but matched correctly according to the semantics of the generated terms and a SYMPTOM term-ID was provided as a final resolution. The term-ID and the label of the best term were manually verified using the Open Lookup Service (OLS) ontology search[21].

## Troubleshooting

During the development and execution of the semantic matching computational steps, several practical issues were encountered. Here we document these issues and provides guidance for resolving them.

1. **Google Cloud Platform notebook timeouts.** Long-running semantic matching tasks may not complete before the 12-hour timeout threshold imposed by Google Cloud Platform. This timeout affects only the Jupyter Notebook interface; the underlying processes continue

Table S1: Minority-vote analysis of 20 generated symptoms, literal ontology matching, and minority-matched SYMP terms, related to Step “Cross-Model Verification”.

| Generated | In Onto? | Minority-Voted Term | Onto-ID |
|---|---|---|---|
| pulmonary hemorrhage | No | bronchopulmonary bleeding | SYMP:0000739 |
| bright red rash | No | blotchy red rash | SYMP:0000743 |
| dry skin | No | skin desquamation | SYMP:0000715 |
| copper dystrophy | No | conjunctival degeneration | SYMP:0020036 |
| progressive neurological deterioration | No | progressive weakness | SYMP:0000363 |
| copper-colored skin | No | icteric skin | SYMP:0000872 |
| unilateral lymphadenopathy | No | lymphadenopathy | SYMP:0019142 |
| frequent pneumothorax | No | pleuritic chest pain | SYMP:0000430 |
| hypotonia | No | motor weakness | SYMP:0000329 |
| infectious mononucleosis-like syndrome | No | pel-epstein fever | SYMP:0000887 |
| hyperkeratotic lesions | No | obsolete hyperkeratotic plaques of surface of foot | SYMP:0000055 |
| copper-colored hue of the skin | No | change in skin color | SYMP:0000184 |
| repeated lung infections | No | lesions in lung | SYMP:0000082 |
| osteopenia | No | obsolete fragile bones | SYMP:0000035 |
| unexplained swelling of lymph nodes | No | lymphadenopathy | SYMP:0019142 |
| copper glycosylation deficit | No | memory impairment | SYMP:0000719 |
| butterfly rash | No | rash | SYMP:0000487 |
| copper-coloredwillappearinjson | No | darkening of skin | SYMP:0000018 |
| copper deposition | No | caseation | SYMP:0020025 |
| loss of mental skills | No | inability to think clearly | SYMP:0000233 |

running on the virtual machine. Users should monitor or terminate these processes through the terminal or an SSH session rather than relying solely on the notebook interface. 685 686

2. **Malformed or partial JSON outputs from LLMs.** Some LLMs occasionally produce incomplete or improperly formatted JSON. To mitigate this, we implemented a robust JSON-extraction and validation step that isolates the `<json>...</json>` block before parsing. This prevents pipeline failures and ensures that only valid JSON is processed. The following code snippet demonstrates the approach: 687 688 689 690 691

```
    text = text.strip()

    # Robust regex to extract JSON block
    match = re.search(r"(?is)<json>\s*(.*?)\s*</json>", text)
    if not match:
        return None

    json_str = match.group(1).strip()

    try:
        return json.loads(json_str)
    except Exception:
        return None
```

3. **Gated model access in vLLM.** Some open-source models require authentication through Hugging Face. Users must log in, generate an access token, and supply it when launching vLLM. Without this token, gated models will fail to load. 692 693 694

4. **Out-of-memory errors when loading large models.** Users working with limited GPU resources may encounter memory errors. Selecting appropriately sized models (e.g., 3B–7B) or reducing batch sizes can prevent these failures. 695 696 697

5. **vLLM server not responding.** The vLLM server may occasionally fail to respond due to port conflicts or stale processes occupying GPU memory. Restarting the server, verifying the port assignment, and ensuring no previous instance is running (e.g., via `nvidia-smi`) typically resolves the issue. 698 699 700 701

# Results 702

The majority vote process resulted in a striking outcome. Whereas 76.85% of the generated terms were voted by the majority vote (i.e. $\rightarrow 4$ votes) resulting in mapping the terms to a precise SYMPTOM ontology ID, the remaining 23.15% of terms were voted by the minority votes demonstrated where at least 2 models agreed on the semantic matching produced by the various LLMs. The minority vote recorded 23.15 which also means that all the generated symptoms were at least matched by 2 votes among 7 models. This process guaranteed that generated terms were authentically associated the SYMPTOM ontology as a credible source symptoms, and assigned a valid Term-ID. Figure S2 shows the outcome of the majority vote, minority vote and the percentage of hallucination of terms that are not symptoms. And, the result is also visualized in Figure S3 703 704 705 706 707 708 709 710 711 712

# Conclusions and Discussions 713

In this work, we distinguish between SYMPTOM ontology exact match against the ontology term label, and the semantic match of the same term using a LLM. The semantic matching here could indicate that the term generated is lexically different from the matching term in the ontology. To 714 715 716

```
print("Majority Counter: ", majority_counter)
print("Minority Counter: ", minority_counter)
print("No Match Counter: ", no_consensus_counter)
print("------------------------------------------------------")
print("Majority    %:", majority_counter/len(majority_results)*100)
print("Minority    %:", round(minority_counter/len(majority_results) * 100, 2))
print("No Consensus %:", no_consensus_counter/len(majority_results)*100)
print("------------------------------------------------------")

# Output:
Majority Count      :  7685
Minority Count      :  2315
No Consensus Count  :  0
----------------------------------------
Majority           %:  76.85
Minority           %:  23.15
No Consensus       %:  0.0
----------------------------------------
```

Figure S2: Summary of majority/minority statistics with corresponding output.

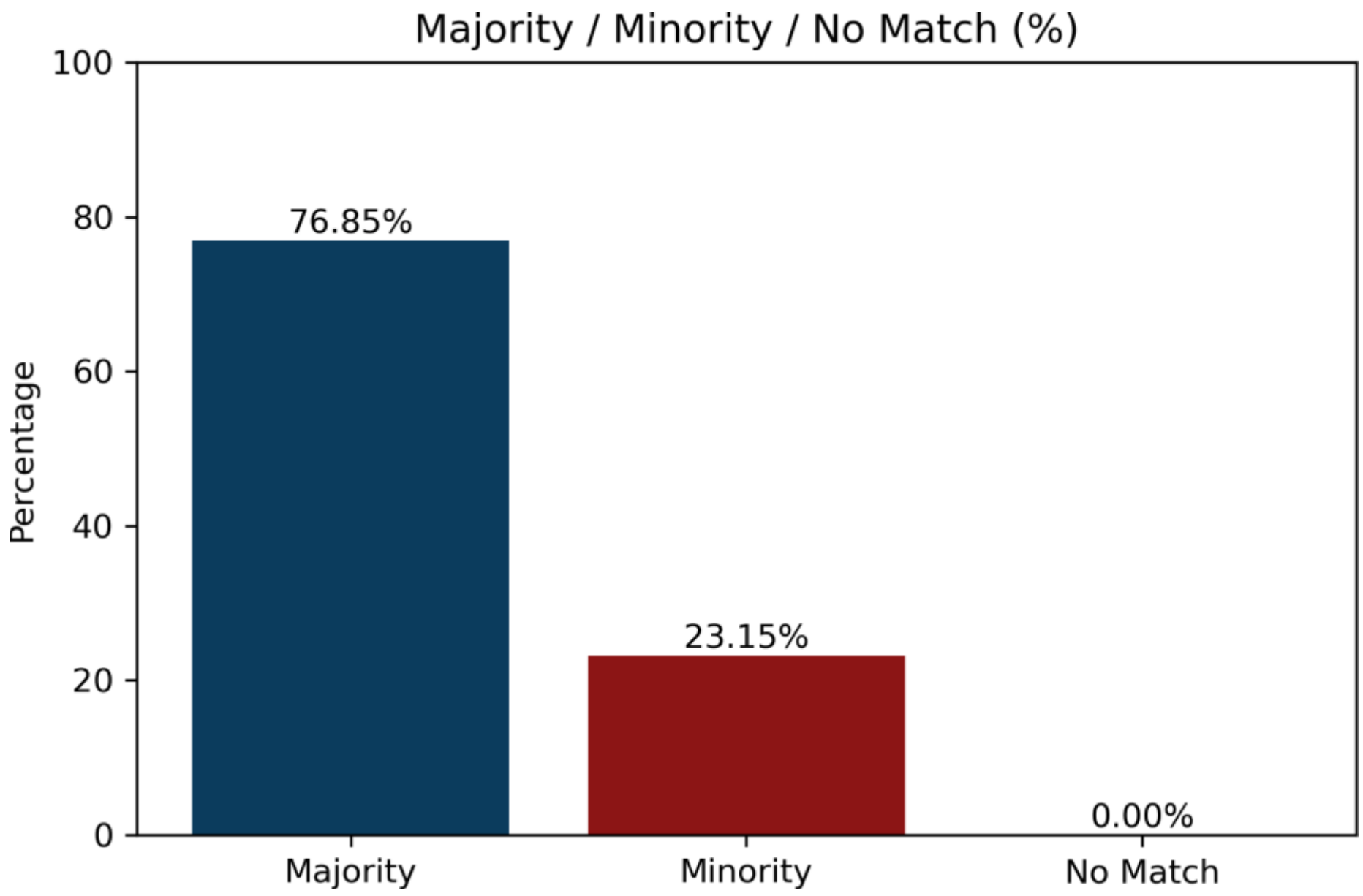

Figure S3: Distribution of majority, minority, and no-match outcomes across the 10,000 generated terms.

demonstrate, an exact match of a generated term is when there is a term in the ontology that has a designated term ID such as “lymphadenopathy” has an exact counterpart and we can identify its Term ID “SYMP:0019142”. As for the semantic match, the generated term would be lexically different. When an LLM model generates a symptom such as “puffy face”, the RAG identifies the top-k candidate terms, and then the LLM executes the prompt to identify the best match, among the k-terms, according to the meaning of the generated term. In this instance, the ontology matching term was identified as ’facial edema’ and the associated term ID “SYMP:0000748”. The cross-model verification and voting process successfully identified semantic matches for all 10,000 generated terms, and no hallucinations were observed in this experiment.

The cross-model verification followed by a majority vote verification process proved that the terms were accurately mapped to precise ontology terms. Compared to the earlier results of using the SYMPTOM ontology exact match, the work promoted the initial work from accuracy of 40%-60% exact match to 100% semantic match agreement across models. More precisely, the match was performed against the ontology terms resulting in all generated terms receiving a semantic mapping through either majority or minority voting. It is important to state that some of the semantically identified terms were voted by $\rightarrow 4$ LLMs which constitutes the majority of the 7 LLMs used. On the other hand, other terms were voted by $\rightarrow 2$ and $< 4$ . Though a minority vote may not be as strong evidence as the majority, it is indeed a much better improvement than having terms that cannot be matched even if it is only by two models. Here we argue that in the event of two independent models agreeing on the semantic match, this provides a basis for verification. This can be achieved by introducing new LLMs to establish a definite majority vote from the newly introduced LLMs until a definite conclusion is made.

This outcome shows the significance of using the cross-model approach presented in this use case and its contribution to the initial study of using ontology term exact match. Among the observations noted, we found that generated terms were matched properly to their exact term, though it was performed semantically and resolved with a proper Term ID. We also found that generated terms that did not have an exact match have been matched to the corresponding ontology term and its exact term ID according to LLMs' built-in similarity capability. The use of LLMs to resolve generated symptom terms and associate them with a precise SYMPTOM ontology is both powerful and promising.

Looking ahead, the strategy presented here can further support the effort of using LLMs to generate knowledge from specific sources and can be verified using the same process here. Particularly, prompting an LLM to generate biomedical terms, from a given text-based document, as part of the prompt will enhance the chances of generating valid terms from the given text. This approach can apply to publications, clinical trials, or any other text document that has medical significance. This process will provide reliable names and relationships, which are a bottleneck for information retrieval tasks [22]. Depending on the features of interest, we may utilize various ontologies (biological, medical, etc). In our future research we will continue to expand on the knowledge learned in this use case using the SYMPTOM ontology and utilize more biomedical ontologies to safeguard against hallucination while we are harnessing the powerful aspects of LLMs[9]. Further, we plan to push the boundaries and evaluate reasoning-focused models. We will utilize the lessons learned from this work and continue to distinguish between what is hallucination and what is real knowledge. It is our goal to develop a general-purpose defensive approach to make the most of LLMs responsibly and discern hallucination from real knowledge.